\documentclass[final]{cvpr}

\usepackage{times}
\usepackage{epsfig}
\usepackage{graphicx}
\usepackage{amsmath}
\usepackage{amssymb}
\usepackage{enumitem}
\usepackage{subcaption}
\usepackage{color}
\usepackage{algorithm}
\usepackage{algorithmic}
\usepackage{diagbox}
\usepackage{array,multirow}

\usepackage[numbers,sort]{natbib} % sorting

\newcolumntype{L}[1]{>{\raggedright\let\newline\\\arraybackslash\hspace{0pt}}m{#1}}
\newcolumntype{R}[1]{>{\raggedleft\let\newline\\\arraybackslash\hspace{0pt}}m{#1}}
\newcolumntype{C}[1]{>{\centering\let\newline\\\arraybackslash\hspace{0pt}}m{#1}}

\usepackage[pagebackref=true,breaklinks=true,colorlinks,bookmarks=false]{hyperref}

\def\cvprPaperID{4631} % *** Enter the CVPR Paper ID here
\def\confYear{CVPR 2021}
\begin{document}

%%%%%%%%% TITLE
\title{Network Quantization with Element-wise Gradient Scaling}

\author{Junghyup Lee \quad\quad\quad Dohyung Kim \quad\quad\quad Bumsub Ham\thanks{Corresponding author.}\vspace*{0.2cm}\\
{School of Electrical and Electronic Engineering, Yonsei University}}
% \and

\maketitle
\thispagestyle{empty}

%%%%%%%%% ABSTRACT
\begin{abstract}
   Network quantization aims at reducing bit-widths of weights and/or activations, particularly important for implementing deep neural networks with limited hardware resources. Most methods use the straight-through estimator~(STE) to train quantized networks, which avoids a zero-gradient problem by replacing a derivative of a discretizer~(i.e., a round function) with that of an identity function. Although quantized networks exploiting the STE have shown decent performance, the STE is sub-optimal in that it simply propagates the same gradient without considering discretization errors between inputs and outputs of the discretizer. In this paper, we propose an element-wise gradient scaling~(EWGS), a simple yet effective alternative to the STE, training a quantized network better than the STE in terms of stability and accuracy. Given a gradient of the discretizer output, EWGS adaptively scales up or down each gradient element, and uses the scaled gradient as the one for the discretizer input to train quantized networks via backpropagation. The scaling is performed depending on both the sign of each gradient element and an error between the continuous input and discrete output of the discretizer. We adjust a scaling factor adaptively using Hessian information of a network. We show extensive experimental results on the image classification datasets, including CIFAR-10 and ImageNet, with diverse network architectures under a wide range of bit-width settings, demonstrating the effectiveness of our method.
   
\end{abstract}

%%%%%%%%% BODY TEXT
\section{Introduction}
Convolutional neural networks~(CNNs) have shown remarkable advances in many computer vision tasks, such as image classification~\cite{krizhevsky2012imagenet,simonyan2014very,he2016deep}, semantic segmentation~\cite{long2015fully,he2017mask}, object detection~\cite{girshick2014rich,liu2016ssd}, and image restoration~\cite{dong2014learning}, while at the cost of large amounts of weights and operations. Network quantization lowers bit-precision of weights and/or activations in a network. It is effective in particular to reduce the memory and computational cost of CNNs, and thus network quantization could be a potential solution for implementing CNNs with limited hardware resources. For example, binarized neural networks~\cite{hubara2016binarized,rastegari2016xnor} use~$32\times$ less memory compared to the full-precision~(32-bit) counterparts, and the binarization techniques allow to replace multiplication and addition with XNOR and bit-count operations, respectively.

\begin{figure}[t]
   \captionsetup{font={small}}
   \begin{center}
      \begin{subfigure}{\columnwidth}
         \centering
         \includegraphics[width=0.8\columnwidth]{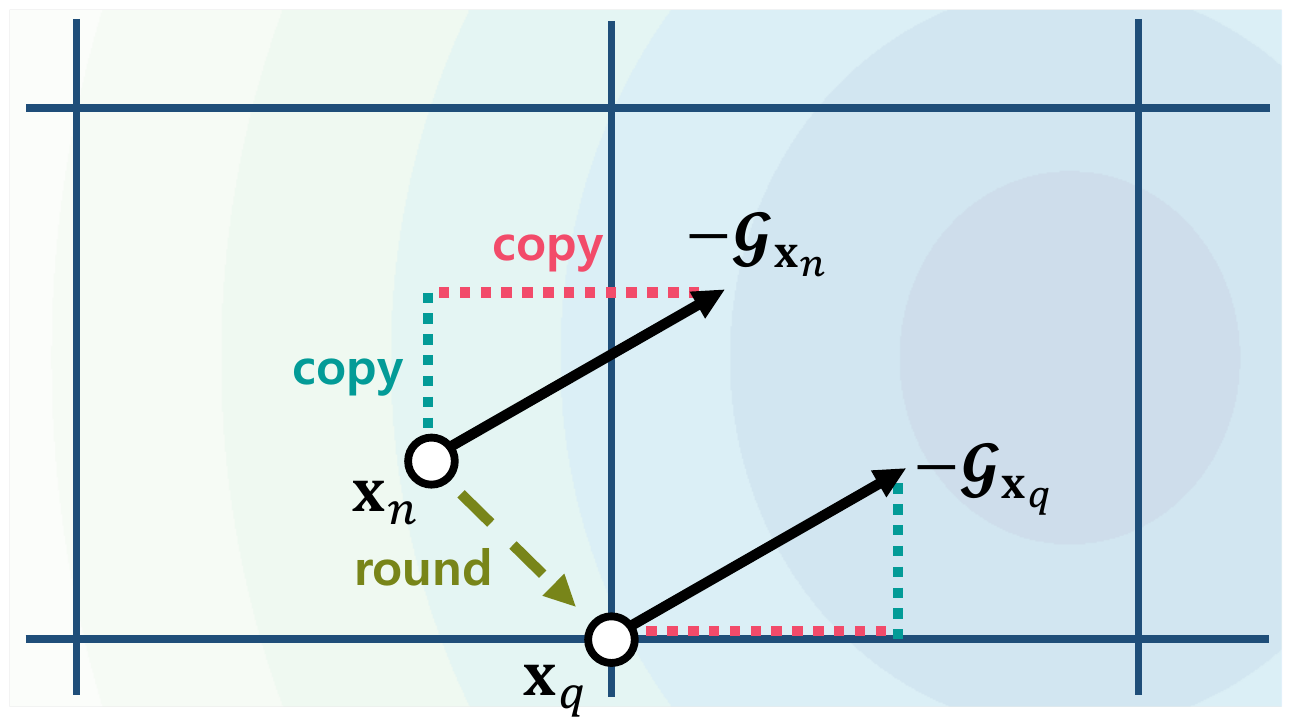}
         \caption{Gradient propagation using STE~\cite{bengio2013estimating}.}
         \label{fig:teaser_STE}
      \end{subfigure}
      \begin{subfigure}{\columnwidth}
         \centering
         \includegraphics[width=0.8\columnwidth]{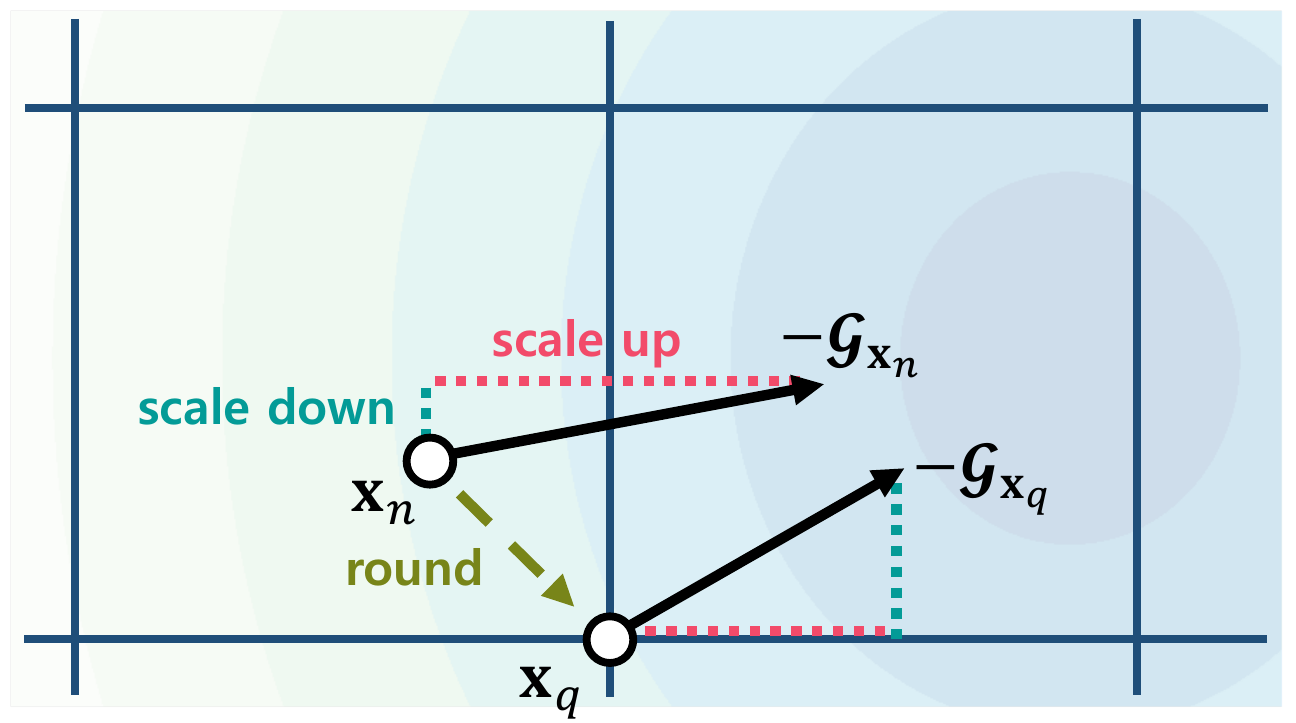}
         \caption{Gradient propagation using EWGS.}
         \label{fig:teaser_ours}
      \end{subfigure}
   \end{center}
   \vspace{-0.5cm}
      \caption{Comparison of STE~\cite{bengio2013estimating} and EWGS. We visualize discrete levels and a loss landscape by straight lines and a contour plot, respectively. In a forward pass, a continuous latent point~$\mathbf{x}_n$ is mapped to a discrete point~$\mathbf{x}_q$ using a round function. Training a quantized network requires backpropagating a gradient from~$\mathbf{x}_q$ to~$\mathbf{x}_n$. (a) The STE propagates the same gradient~\ie,~$\mathcal{G}_{\mathbf{x}_n}=\mathcal{G}_{\mathbf{x}_q}$ without considering the value of~$\mathbf{x}_n$, where we denote by~$\mathcal{G}_{\mathbf{x}_n}$ and~$\mathcal{G}_{\mathbf{x}_q}$ the gradients of~$\mathbf{x}_n$ and~$\mathbf{x}_q$, respectively. (b) Our approach, on the other hand, scales up or down each element of the gradient during backpropagation, while taking into account discretization errors~\ie,~${\mathbf{x}_n} - {\mathbf{x}_q}$. (Best viewed in color.)}
   \vspace{-0.45cm}
   \label{fig:teaser}
\end{figure}

Quantized networks involve weight and/or activation quantizers in convolutional or fully-connected layers. The quantizers take full-precision weights or activations, and typically perform normalization, discretization, and denormalization steps to convert them into low-precision ones. The main difficulty of training a quantized network arises from the discretization step, where a discretizer~(\ie, a round function) maps a normalized value to one of discrete levels. Since an exact derivative of the discretizer is either zero or infinite, gradients become zero or explode during backpropagation. Most quantization methods~\cite{zhou2016dorefa,choi2018pact,jung2019learning,zhang2018lq,esser2019learned,park2020profit} overcome this issue by exploiting the straight-through estimator~(STE)~\cite{bengio2013estimating}. The STE propagates the same gradient from an output to an input of the discretizer, assuming that the derivative of the discretizer is equal to~$1$. This could bring a gradient mismatch problem~\cite{yang2019quantization}, since the discretizer used in a forward pass~(\ie, the round function) does not match up with that in a backward pass~(\ie, an identity or hard tanh functions). Nevertheless, recent methods exploiting the STE have shown reasonable performance~\cite{jung2019learning,esser2019learned,bhalgat2020lsq+,park2020profit}.

We take a different point of view on how the STE works. We interpret that a full-precision input~(which we call a ``latent value'') of the discretizer moves in a continuous space, and a discretizer output~(which we call a ``discrete value'') is determined by projecting the latent value to the nearest discrete level in the space. This suggests that shifting the latent values in the continuous space influences the discrete values. The STE, in this sense, shifts~(or updates) the latent values with \emph{coarse gradients}~\cite{yin2019understanding}, that is, the gradients obtained with the discrete values~(Fig.~\ref{fig:teaser_STE}), which is sub-optimal. For example, both latent values of~$0.51$ and~$1.49$ produce the same discrete value of~$1$ using a round function, and the STE forces to update the latent values equally with the same gradient from the discrete value of~$1$, regardless of their discretization errors induced by the rounding. Updating these latent values should be treated differently, because, for example, a small increment for the latent value of~$1.49$ leads to changing the discrete value from~$1$ to~$2$, whereas the increment for the latent value of~$0.51$ cannot. Similarly, a small decrement for the latent value of~$0.51$ can convert the discrete value from~$1$ to~$0$, but the latent value of~$1.49$ requires a much larger decrement to do so.

In this paper, we present an element-wise gradient scaling~(EWGS) that enables better training of a quantized network, compared with the STE, in terms of stability and accuracy. Given a gradient of discrete values, EWGS adaptively scales up or down each element of the gradient considering its sign and discretization errors between latent and discrete values. The scaled gradient is then used to update the latent value~(Fig.~\ref{fig:teaser_ours}). Since optimal scaling factors, which control the extent of EWGS, may vary across weight or activation quantizers in different layers, we propose an approach to adjusting the factors adaptively during training. Specifically, we relate the scaling factor with the second-order derivatives of a task loss w.r.t the discrete values, and propose to estimate the factor with the trace of a Hessian matrix, which can be computed efficiently with the Hutchinson's method~\cite{avron2011randomized,yao2019pyhessian}. Without an extensive hyperparameter search, training schedules~\cite{zhou2017incremental,zhuang2018towards,yang2019quantization}, or additional modules~\cite{mishra2017apprentice,zhuang2020training,chen2019metaquant}, various CNN architectures trained with our approach achieve state-of-the-art performance on ImageNet~\cite{deng2009imagenet}. Note that the STE is a special case of EWGS, indicating that it can be exploited to other quantization methods using the STE. The main contributions of our work can be summarized as follows:
\begin{itemize}[leftmargin=*]
   \item[$\bullet$] We introduce EWGS that scales up or down each gradient element of the discrete value adaptively for backpropagation, while considering discretization errors between inputs and outputs of a discretizer.
   \item[$\bullet$] We relate a scaling factor with the second-order derivatives of a loss function w.r.t discrete values, allowing to compute the parameter effectively and adaptively with the Hessian information of a quantized network.
   \item[$\bullet$] We demonstrate the effectiveness of our method with various CNN architectures under a wide range of bit-widths, outperforming the state of the art on ImageNet~\cite{deng2009imagenet}. We also verify that our approach boosts the performance of other quantization methods, such as DoReFa-Net~\cite{zhou2016dorefa} and PROFIT~\cite{park2020profit}.
\end{itemize} 
Our code and models are available online:~\url{https://cvlab.yonsei.ac.kr/projects/EWGS}.

\section{Related work}
Network quantization has been formulated as a constrained optimization problem to minimize quantization errors, where bit-widths of weight and/or activation values are restricted by binary~\cite{rastegari2016xnor}, ternary~\cite{li2016ternary,zhu2016trained}, or arbitrary ones~\cite{zhou2016dorefa}. The works of~\cite{cai2017deep,wang2018two} propose to consider the half-wave Gaussian distribution of activations for quantization, resulting from batch normalization~\cite{ioffe2015batch} and a ReLU~\cite{krizhevsky2012imagenet}, which reduces the errors from quantizing activation values. Recent methods learn quantizer parameters for controlling, \eg, clipping ranges~\cite{choi2018pact,jung2019learning,esser2019learned} and non-uniform quantization intervals~\cite{jung2019learning} or levels~\cite{zhang2018lq}. Motivated by this, we design a uniform quantizer, and learn lower and upper bounds of quantization intervals~\cite{jung2019learning}. All the aforementioned approaches exploit STE to handle the derivative of a discretizer. This suggests that our approach can be easily incorporated into these methods, making it possible to boost the performance in a complementary way. 

Aside from quantization methods, lots of training techniques have been introduced to enhance the performance of quantized networks. Incremental quantization~\cite{zhou2017incremental} divides network weights in a layer into two groups of full-precision and quantized ones, and trains a quantized network in an iterative manner by expanding the group of quantized weights gradually. Progressive quantization~\cite{zhuang2018towards} decreases bit-widths from high- to low-precision gradually, boosting the performance of a low-precision model. To leverage the knowledge from full-precision models, high-performance networks~\cite{mishra2017apprentice} or layer-wise auxiliary modules~\cite{zhuang2020training} are also exploited. Very recently, PROFIT~\cite{park2020profit} introduces a training strategy, specially designed for quantizing light-weight networks, which progressively freezes learned network weights during an iterative quantization process. These methods also rely on STE, and require heuristic scheduling techniques~\cite{zhou2017incremental,zhuang2018towards,park2020profit} or additional network weights~\cite{mishra2017apprentice,zhuang2020training} for training. In contrast, our approach focuses on the backpropagation step of network quantization, improving the performance without bells and whistles.

Similar to ours, recent methods~\cite{yang2019quantization,gong2019differentiable,bai2018proxquant,chen2019metaquant} try to tackle the problem of STE. They claim that the STE causes a gradient mismatch problem~\cite{yang2019quantization,gong2019differentiable}, and introduce soft versions of discretizers, consisting of sigmoid~\cite{yang2019quantization} or tanh~\cite{gong2019differentiable}. These approaches approximate the discretizer~(typically using a round function) well, especially when a temperature parameter in sigmoid or tanh functions is large, but the large temperature causes vanishing/exploding gradient problems. Hyperparameters should thus be tuned carefully during training~\cite{yang2019quantization}. The proximal method with a regularizer~\cite{bai2018proxquant} and a meta quantizer using synthetic gradients~\cite{chen2019metaquant} avoid the use of STE. They are, however, limited to quantizing network weights only, and require a cost-expensive optimization process~\cite{bai2018proxquant} or additional meta-learning modules~\cite{chen2019metaquant}. On the contrary, our method can be applied to both weight and activation quantization in an efficient manner with a simple gradient scaling.

Closely related to ours, the works of~\cite{dong2019hawq,dong2019hawqv2} exploit Hessian information for network quantization. Specifically, they exploit eigenvalues~\cite{dong2019hawq} or traces~\cite{dong2019hawqv2} of Hessian matrices to measure the sensitivity of each layer, and allocate different bit-widths to the layers. That is, they leverage the Hessian information to train a mixed-precision network. In contrast to this, we use the trace of a Hessian matrix to adjust a scaling factor for EWGS.

\begin{figure*}[t]
   \captionsetup{font={small}}
   \begin{center}
      \begin{subfigure}{\textwidth}
         \centering
         \hspace{0.75cm}
         \includegraphics[width=0.9\textwidth]{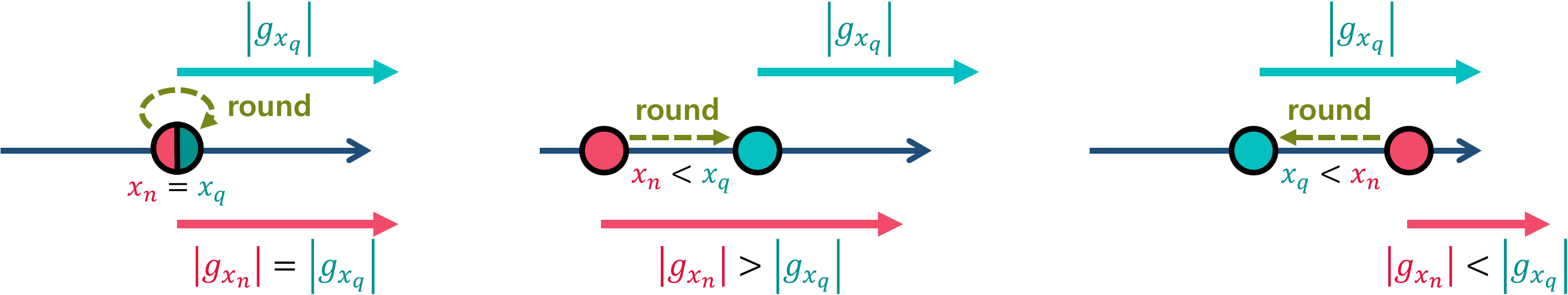}
         \caption{The sign of an update for the discrete value~$x_q$ is positive~(\ie, $-g_{x_q} > 0$).}
         \vspace{0.5cm}
         \label{fig:method_pos}
      \end{subfigure}
      \begin{subfigure}{\textwidth}
         \centering
         \hspace{-0.4cm}
         \includegraphics[width=0.875\textwidth]{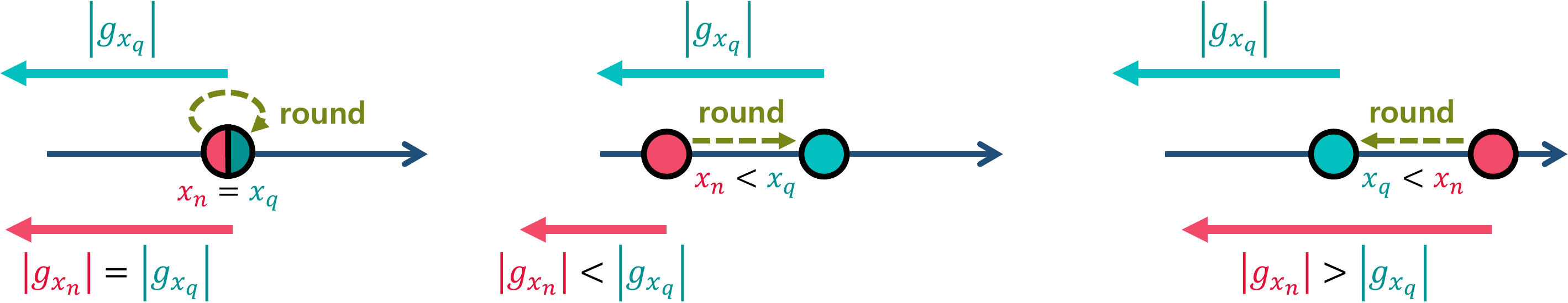}
         \caption{The sign of an update for the discrete value~$x_q$ is negative~(\ie, $-g_{x_q} < 0$).}
         \label{fig:method_neg}
      \end{subfigure}
   \end{center}
   \vspace{-0.45cm}
      \caption{1-D illustrations of EWGS. We visualize a latent value~$x_n$ and a discrete value~$x_q$, by red and cyan circles, respectively, where the discrete value is obtained by applying a round function~(a dashed arrow) to the latent value. We also visualize their update vectors by solid arrows with corresponding colors, and we denote by~$\vert g_{x_n} \vert$ and~$\vert g_{x_q} \vert$ the magnitudes of the update vectors for~$x_n$ and~$x_q$, respectively. For each (a) and (b), we present three cases, where the latent value~$x_n$ is equal to~(left), smaller than~(middle), and larger than~(right) the discrete one~$x_q$. EWGS scales up the gradient element for the discrete value~$g_{x_q}$, when a latent value~$x_n$ requires a larger magnitude of an update, compared to the discrete one~$x_q$~(\eg, (a)-middle or (b)-right), and scaling down in the opposite case~(\eg, (a)-right or (b)-middle). When the latent value~$x_n$ is equal to the discrete value~$x_q$, it propagates the same gradient element similar to STE~(\eg, (a)-left or (b)-left). (Best viewed in color.)}
   \vspace{-0.4cm}
   \label{fig:method}
\end{figure*}

\section{Approach}
In this section, we introduce our quantization method using EWGS~(Sec.~\ref{sec:grad_scaling}). We then describe how to determine a scaling factor for EWGS~(Sec.~\ref{sec:scaling_factor}).

\subsection{Quantization with EWGS} \label{sec:grad_scaling}
We design a uniform quantizer~$Q$ that converts a full-precision input~$x$ to a quantized output~$Q({x})$, where~we denote by $x$ a scalar element of either a weight or an input activation tensor~$\mathbf{x}$ in a layer. We learn a quantization interval~\cite{jung2019learning,gong2019differentiable} using lower and upper bounds, denoted by $l$ and $u$, respectively. Specifically, the quantizer first generates a full-precision latent value~$x_{n}$ by normalizing and clipping the input value~$x$ as follows:
\begin{equation} \label{eq:latent_value}
   x_{n} = \textrm{clip}\left(\frac{x-l}{u-l}, 0, 1\right),
\end{equation}
where $\textrm{clip}(\cdot,0,1)$ is a clipping function with lower and upper bounds of~$0$ and~$1$, respectively. Note that weight and/or activation quantizers in every quantized layer use separate parameters for the quantization intervals~(\ie, $l$ and~$u$). For~$b$-bit quantization, the latent value~$x_{n}$ is converted to a discrete value~$x_{q}$ using a round function with pre-/post-scaling as follows:
\begin{equation} \label{eq:round}
   x_{q} = \frac{\textrm{round}((2^{b}-1)x_{n})}{2^{b}-1}.
\end{equation}
Finally, the quantizer outputs a quantized weight~$Q_{W}(x)$ or activation~$Q_{A}(x)$ as follows:
\begin{equation} \label{eq:quantize}
   Q_{W}(x) = 2\left(x_{q} - 0.5\right),~Q_{A}(x) = x_{q},
\end{equation}
where we restrict the quantized activation~$Q_{A}(x)$ to be non-negative~\cite{cai2017deep} considering the pre-activation by a ReLU. To adjust an output scale of the layer, we train an additional parameter~$\alpha$ for each quantized layer, which is multiplied by the output activations of convolutional or fully-connected layers.

The main difficulty of training a quantized network arises from the round function in Eq.~\eqref{eq:round}, since its derivative is zero at almost everywhere. Most quantization methods avoid zero gradients using STE~\cite{bengio2013estimating}. It approximates the derivative of the round function by an identity function, that is,~$\mathcal{G}_{\mathbf{x}_{n}} = \mathcal{G}_{\mathbf{x}_{q}}$, where we denote by~$\mathbf{x}_{n}$ and~$\mathbf{x}_{q}$~tensors containing latent and discrete values, respectively, and by~$\mathcal{G}_{\mathbf{x}_{n}}$ and~$\mathcal{G}_{\mathbf{x}_{q}}$ corresponding gradients. Propagating the same gradient from discrete to latent values is, however, sub-optimal for the following reasons: (1)~Multiple latent values can produce the same discrete value; (2)~The same gradient provided by the discrete value affects differently to each of the latent values. To overcome this problem, we introduce an EWGS method, an effective alternative to STE, defined as follows:
\begin{equation} \label{eq:grad_scaling}
   g_{x_{n}} = g_{x_{q}} (1 + \delta \mathrm{sign}(g_{x_{q}})(x_{n} - x_{q})),
\end{equation}
where~$g_{x_{n}}$ and~$g_{x_{q}}$ are the elements of the gradients~$\mathcal{G}_{\mathbf{x}_{n}}$ and~$\mathcal{G}_{\mathbf{x}_{q}}$, corresponding to the partial derivatives of a task loss w.r.t $x_{n}$ and $x_{q}$, respectively. $\mathrm{sign}(\cdot)$ is a signum function and $\delta \ge 0$ is a scaling factor. EWGS adjusts the gradient element of discrete values~$g_{x_{q}}$ adaptively using the sign of the element,~$\mathrm{sign}(g_{x_{q}})$, and a discretization error, $x_{n} - x_{q}$. Note that STE is a special case of EWGS, that is, Eq.~\eqref{eq:grad_scaling} corresponds to STE, when a scaling factor~$\delta$ is zero. 

We visualize in Fig.~\ref{fig:method} 1-D examples illustrating an effect of EWGS. We can see that EWGS encourages a gradient element for the discrete value~$g_{x_{q}}$ to decrease with the scale of $(1 + \delta \mathrm{sign}(g_{x_{q}})(x_{n} - x_{q}))$ to update the latent value~$x_{n}$, when the latent value~$x_{n}$ to update is already located farther than the discrete value~$x_{q}$ in the direction of change~(\ie, $-\mathrm{sign}(g_{x_{q}})$) as shown in Fig.~\ref{fig:method_pos} (right) and Fig.~\ref{fig:method_neg} (middle), and to increase in the opposite case as shown in~Fig.~\ref{fig:method_pos} (middle) and Fig.~\ref{fig:method_neg} (right). Note that we use non-negative values for the scaling factor~$\delta$, since negative ones lead to opposite effects. To sum up, EWGS resolves discrepancies between latent and discrete values during backpropagation by considering discretization errors between these values and their direction of change. As will be shown in Sec.~\ref{sec:exp_analy}, this not only stabilizes the training of a quantized network but also encourages better convergence compared to STE.

\begin{algorithm}[t]
	\caption{Forward and backward propagations in a quantizer using EWGS.} 
	\label{alg:algorithm}
	\small
	\begin{algorithmic}[1]
      \STATE \textbf{Hyperparameter}: a quantization bit-width~$b$; an update period of the scaling factor~$k$.

      \STATE \textbf{Parameter}: lower and upper bounds of a quantization interval, denoted by~$l, u \in \mathbb{R}$, respectively; a scaling factor for EWGS~$\delta \in \mathbb{R}$.

      \STATE \textbf{Input}: a full-precision input tensor~$\mathbf{x} \in \mathbb{R}^{N}$ containing either weights or activations, where~$N$ is the number of elements in the tensor.

      \STATE \textbf{Output}: a quantized tensor~$Q(\mathbf{x}) \in \mathbb{R}^{N}$.

		\STATE {\textbf{Forward Propagation}}
      \STATE Compute latent values [Eq.~\eqref{eq:latent_value}]:\\
      $\mathbf{x}_n = \textrm{clip} \left( \frac{\mathbf{x} - l}{u - l}, 0, 1 \right)$.\\
		\STATE Compute discrete values [Eq.~\eqref{eq:round}]:\\
      $\mathbf{x}_q = \frac{\textrm{round}((2^b-1)\mathbf{x}_n)}{2^b-1}$.\\ 
      \STATE Compute quantized output values [Eq.~\eqref{eq:quantize}]:\\
      $
         Q(\mathbf{x})=
         \begin{cases}
            2\left(\mathbf{x}_{q} - 0.5\right), & \text{if } \mathbf{x} \text{ is a weight tensor.} \\
            \mathbf{x}_{q}, & \text{if } \mathbf{x} \text{ is an activation tensor.}
         \end{cases}
       $

      \STATE {\textbf{Backward Propagation}}
      \STATE Obtain the gradient of discrete values~$\mathcal{G}_{\mathbf{x}_{q}}$ via backpropagation.\\
      \STATE Calculate the gradient of latent values~$\mathcal{G}_{\mathbf{x}_{n}}$ using EWGS [Eq.~(\ref{eq:grad_scaling})]:\\
      $\mathcal{G}_{\mathbf{x}_{n}}=\mathcal{G}_{\mathbf{x}_{q}} \odot \left( 1 + \delta \mathrm{sign}(\mathcal{G}_{\mathbf{x}_{q}}) \odot (\mathbf{x}_n-\mathbf{x}_q) \right)$, \\
      where~$\odot$ is element-wise multiplication and $\mathrm{sign}(\cdot)$ applies the signum function to each element.
      \STATE Propagate the gradient to the input.

      \STATE {\textbf{Scaling Factor Update}}
      \STATE Update the scaling factor~$\delta$ using Eq.~\eqref{eq:update_scale} for every~$k$ iterations.
   \end{algorithmic}
\end{algorithm}

\subsection{Scaling factor for EWGS} \label{sec:scaling_factor}
It is crucial to determine a scaling factor~$\delta$, since an improper value would hinder the training process, and weight/activation quantizers in different layers may require different degrees of scaling. Let us consider the following equation:
\begin{equation} \label{eq:derivation1}
   \begin{split}
   g_{x_{n}} &= g_{x_{q}} + \frac{g_{x_{n}} - g_{x_{q}}}{x_{n} - x_{q}} \left(x_{n}-x_{q}\right) \\
             &= g_{x_{q}} + \frac{g_{x_{q}+\epsilon} - g_{x_{q}}}{\epsilon} \left(x_{n}-x_{q}\right),
   \end{split}
\end{equation}
where~$\epsilon = x_{n} - x_{q}$ is a discretization error from the round function in Eq.~\eqref{eq:round}.
Since an absolute value of the error is bounded by a small number \ie,~$\vert \epsilon \vert \le \frac{0.5}{2^b-1}$, we assume that the error is small enough to approximate Eq.~\eqref{eq:derivation1} as follows:
\begin{equation} \label{eq:derivation2}
   g_{x_{n}} \approx g_{x_{q}} + g_{x_{q}}^{\prime} \left(x_{n}-x_{q}\right),
\end{equation}
where~$g_{x_{q}}^{\prime}=\frac{\partial g_{x_{q}}}{\partial x_{q}}$ is a second-order derivative of a task loss w.r.t the discrete value~$x_{q}$. This can be represented as follows:
\begin{equation} \label{eq:derivative3}
   g_{x_{n}} \approx g_{x_{q}} \left(1 + \frac{g_{x_{q}}^{\prime}}{\vert g_{x_{q}} \vert} \mathrm{sign}(g_{x_{q}}) (x_{n}-x_{q}) \right),
\end{equation}
which corresponds to EWGS in Eq.~\eqref{eq:grad_scaling}. This suggests that we can set the scaling factor~$\delta$ as~$\frac{g_{x_{q}}^{\prime}}{\vert g_{x_{q}} \vert}$, but calculating an exact Hessian matrix~$H$ to obtain the second-order derivative~$g_{x_{q}}^{\prime}$ is computationally demanding. We instead approximate the second-order derivative by an average of diagonal elements in the Hessian matrix~$H$, with an assumption that the main diagonal dominates the matrix~$H$, and the discrete values~$\mathbf{x}_{q}$ obtained from the same weight or activation quantizer in a layer influence similarly to the loss function~\cite{dong2019hawq,dong2019hawqv2}. To this end, we compute a Hessian trace with an efficient algorithm~\cite{dong2019hawqv2,yao2019pyhessian} using the Hutchinson's method~\cite{avron2011randomized}:
\begin{equation} \label{eq:hutchinson}
   \begin{split}
   \mathrm{Tr}(H) &= \mathrm{Tr}(HI) = \mathrm{Tr}(H\mathbb{E}[\mathbf{vv}^T]) \\
                  &= \mathbb{E}[\mathrm{Tr}(H\mathbf{vv}^T)] = \mathbb{E}[\mathbf{v}^TH\mathbf{v}],
   \end{split}
\end{equation}
where~$I$ is an identity matrix,~$\mathbb{E}$ is an expectation operator, and~$\mathbf{v}$ is a random vector drawn from the Rademacher distribution, satisfying~$\mathbb{E}[\mathbf{vv}^T]=I$. This implies that we can estimate the trace of a Hessian matrix~$\mathrm{Tr}(H)$ with~$\mathbb{E}[\mathbf{v}^TH\mathbf{v}]$, where we can obtain~$H\mathbf{v}$ efficiently without forming an exact Hessian matrix as follows:
\begin{equation}
   \frac{\partial{\mathcal{G}_{\mathbf{x}_{q}}^T\mathbf{v}}}{\partial{\mathbf{x}_{q}}} = \frac{\partial{\mathcal{G}_{\mathbf{x}_{q}}^T}}{\partial{\mathbf{x}_{q}}}\mathbf{v} + \mathcal{G}_{\mathbf{x}_{q}}^T \frac{\partial{\mathbf{v}}}{\partial{\mathbf{x}_{q}}} 
   = \frac{\partial{\mathcal{G}_{\mathbf{x}_{q}}^T}}{\partial{\mathbf{x}_{q}}}\mathbf{v} = H\mathbf{v}.
\end{equation}
We then define the scale parameter~$\delta$ as follows:
\begin{equation} \label{eq:update_scale}
   \delta = \frac{\mathrm{Tr}(H)/N}{G},
\end{equation}
where~$N$ is the number of diagonal elements in the Hessian matrix and~$G$ is a gradient representative determined from a distribution of the gradients~$\mathcal{G}_{\mathbf{x}_{q}}$. Based on Eq.~\eqref{eq:derivative3}, we could take an average over the absolute values of gradient elements~\ie,~$\mathbb{E}[\vert g_{x_q} \vert]$ for setting~$G$, but we empirically found that most gradients are concentrated near zero, such that the average value tends to be biased to small gradient elements. We instead set~$G$ to a sufficiently large value. A plausible reason is that considering large gradient elements is more important, since they dominate the training. Specifically, we use~$3\sigma(\mathcal{G}_{\mathbf{x}_{q}})$ as the gradient representative~$G$, where $\sigma(\cdot)$ computes a standard deviation. It enables finding a sufficiently large gradient element~(\eg, $3\sigma(\cdot)$ accounts for roughly 99 percent of data in case of the Gaussian distribution). We could take the maximum over the absolute values of gradient elements, but it often corresponds to an outlier of a distribution. 

Assuming that the loss function is locally convex, we take a non-negative value for the scaling factor~(\ie, $\textrm{max}(0, \delta)$), which coincides with the condition in Eq.~\eqref{eq:grad_scaling}. We use individual scaling factors for all weight and activation quantizers in a network, and update them periodically during training for efficiency. We summarize in Algorithm~\ref{alg:algorithm} an overall quantization procedure of our approach.

\section{Experiments}
In this section, we describe our experimental settings~(Sec.~\ref{sec:exp_setting}) and evaluate our method on image classification~(Sec.~\ref{sec:exp_results}). We then present a detailed analysis on EWGS~(Sec.~\ref{sec:exp_analy}).

\subsection{Experimental settings} \label{sec:exp_setting}
\paragraph{Dataset.}
We perform extensive experiments on standard benchmarks for image classification, including CIFAR-10~\cite{krizhevsky2009learning} and ImageNet~(ILSVRC-2012)~\cite{deng2009imagenet}. The CIFAR-10 dataset contains images of size~$32\times32$, consisting of 50K training and 10K test images with 10 classes. The ImageNet dataset includes roughly 1.2M training and 50K validation images with 1K classes. We report the top-1 classification accuracy for both datasets.

\vspace{-0.3cm}
\paragraph{Network architectures.}
We use network architectures of ResNet-20~\cite{he2016deep} on CIFAR-10, and ResNet-18, ResNet-34 and MobileNet-V2~\cite{sandler2018mobilenetv2} on ImageNet. We do not modify the network architectures for fair comparison. We insert weight and/or activation quantizers right before the convolutional or fully-connected operators in every layer to quantize. Following the standard experimental protocol in~\cite{rastegari2016xnor,zhang2018lq}, we do not quantize the first and the last layers unless otherwise specified. We initialize network weights with pretrained full-precision models, which are readily available in PyTorch~\cite{paszke2017automatic}~(ResNet-18, ResNet-34, and MobileNet-V2) or trained by ourselves~(ResNet-20).

\begin{table*}[t]
   \setlength{\tabcolsep}{0.3em}
   \captionsetup{font={small}}
   \centering
   \small
   \begin{tabular}{l|C{1.65cm}|C{1.65cm}|C{1.65cm}|C{1.65cm}|C{1.65cm}|C{1.65cm}|C{1.65cm}||C{1.1cm}}
      \hline
      \diagbox{Methods}{W/A} & 1/1 & 1/2 & 2/2 & 3/3 & 4/4 & 1/32 & 2/32 & 32/32 \\
      \hline
      \hline
      XNOR~\cite{rastegari2016xnor}         & 51.2~($-$18.1)         & -                     & -                     & -                     & -                                      & 60.8~($-$8.5)         & -                     & 69.3 \\
      PACT~\cite{choi2018pact}              & -                      & -                     & 64.4~($-$5.8)         & 68.1~($-$2.1)         & 69.2~~\hspace{0.048cm}($-$1.0)         & -                     & -                     & 70.2 \\
      LQ-Net~\cite{zhang2018lq}             & -                      & 62.6~($-$7.7)         & 64.9~($-$5.4)         & 68.2~($-$2.1)         & 69.3~~\hspace{0.048cm}($-$1.0)         & -                     & 68.0~($-$2.3)         & \bf{70.3} \\
      QIL~\cite{jung2019learning}           & -                      & -                     & 65.7~($-$4.5)         & 69.2~($-$1.0)         & 70.1~~\hspace{0.048cm}($-$0.1)         & -                     & 68.1~($-$2.1)         & 70.2 \\
      QuantNet~\cite{yang2019quantization}  & 53.6~($-$16.7)         & 63.4~($-$6.9)         & -                     & -                     & -                                      & 66.5~($-$3.8)         & 69.1~($-$1.2)         & \bf{70.3} \\
      DSQ~\cite{gong2019differentiable}     & -                      & -                     & 65.2~($-$4.7)         & 68.7~($-$1.2)         & 69.6$^\dagger$~($-$0.3)                & 63.7~($-$6.2)         & -                     & 69.9 \\
      LSQ~\cite{esser2019learned,bhalgat2020lsq+} & -                & -                     & 66.7~($-$3.4)         & 69.4~($-$0.7)         & 70.7~~\hspace{0.048cm}($+$0.6)         & -                     & -                     & 70.1 \\
      LSQ+~\cite{bhalgat2020lsq+}           & -                      & -                     & 66.8~($-$3.3)         & 69.3~($-$0.8)         &\bf{70.8}~~\hspace{0.048cm}(\bf{$+$0.7})& -                     & -                     & 70.1 \\
      IRNet~\cite{qin2020forward}           & -                      & -                     & -                     & -                     & -                                      & 66.5~($-$3.1)         & -                     & 69.6 \\
      Ours                                  &\bf{55.3}~(\bf{$-$14.6})&\bf{64.4}~(\bf{$-$5.5})&\bf{67.0}~(\bf{$-$2.9})&\bf{69.7}~(\bf{$-$0.2})& 70.6~~\hspace{0.048cm}(\bf{$+$0.7})    &\bf{67.3}~(\bf{$-$2.6})&\bf{69.6}~(\bf{$-$0.3})& 69.9 \\
      %                                       1/1                     1/2                      2/2                     3/3                     4/4                                      1/32                   2/32                    32/32
      \hline
   \end{tabular}
   \vspace{-0.2cm}
   \caption{Quantitative comparison of top-1 validation accuracy on ImageNet~\cite{deng2009imagenet} using the ResNet-18~\cite{he2016deep} architecture. We report results for quantized networks and their full-precision versions. W/A represents bit-widths of weights~(W) and activations~(A). The numbers in brackets indicate the performance drops or gains compared to the full-precision models. $^\dagger$: all layers including the first and the last layers are quantized.}
   \label{tab:resnet18}
   \vspace{-0.5cm}
\end{table*}

\vspace{-0.3cm}
\paragraph{Initialization.}
We initialize the lower and upper bounds of a quantization interval,~$l$ and $u$, respectively, by considering the distribution of quantizer inputs, such that the interval covers roughly 99\% of the input values, to use a set of discrete levels effectively. Specifically, for each weight quantizer, the lower and upper bounds are initialized by~$-3\sigma(\mathbf{w})$ and~$3\sigma(\mathbf{w})$, respectively, where~$\mathbf{w}$ is a weight tensor in a layer. Considering that input activations typically follow the half-wave Gaussian distribution~\cite{cai2017deep}, we initialize the lower and upper bounds in each activation quantizer with 0 and~$\frac{3\sigma(\mathbf{a})}{\sqrt{1-2/\pi}}$, respectively, where~$\mathbf{a}$ is an input activation tensor. An output scale~$\alpha$ in every quantized layer is initialized by~$\frac{\mathbb{E}(\vert o \vert)}{\mathbb{E}(\vert o_q \vert)}$, where~$o$ and~$o_q$ are convolution~(or matrix multiplication) outputs computed with full-precision and quantized representations, respectively. Note that we initialize these parameters during the first forward pass. We set scaling factors~$\delta$ for EWGS to 0 initially, and update them for every 1 epoch on ImageNet and 10 epochs on CIFAR-10.
   
\vspace{-0.3cm}
\paragraph{Training details.}
Initial learning rates for network weights are set to 1e-3, 1e-2, 1e-2, and 5e-3 for ResNet-20, ResNet-18, ResNet-34, and MobileNet-V2, respectively. We set a learning rate for the quantizer parameters~(\ie, interval parameters,~$l$ and~$u$, and output scales~$\alpha$) to 1e-5, smaller than those of the network weights~\cite{jung2019learning}. We use a cosine annealing technique~\cite{loshchilov2016sgdr} for learning rate decay. Following the training settings in~\cite{jung2019learning,qin2020forward,park2020profit}, we use the SGD optimizer to train the network weights, except ResNet-20 on CIFAR-10 using the Adam optimizer~\cite{kingma2014adam}, with weight decay of 4e-5 for MobileNet-V2 and 1e-4 for the others. The quantizer parameters are trained with the Adam optimizer without weight decay. We train ResNet-20 for 400 epochs on CIFAR-10 with a batch size of 256. ResNet-18, ResNet-34, and MobileNet-V2 are trained for 100 epochs on ImageNet with batch sizes of 256, 256, and 100, respectively.

\begin{table*}[t]
   \setlength{\tabcolsep}{0.3em}
   \captionsetup{font={small}}
   \centering
   \small
   \begin{tabular}{l|C{1.65cm}|C{1.65cm}|C{1.65cm}|C{1.65cm}|C{1.65cm}|C{1.65cm}||C{1.1cm}}
      \hline
      \diagbox{Methods}{W/A} & 1/1 & 1/2 & 2/2 & 3/3 & 4/4 & 1/32 & 32/32 \\
      \hline
      \hline
      LSQ~\cite{esser2019learned}           & -                       & -                      & 71.6~($-$2.5)          & 73.4~($-$0.7)          & 74.1~~\hspace{0.048cm}($+$0.0)          & -                       & 74.1 \\
      \hline
      ABC-Net~\cite{lin2017towards}         & 52.4~($-$20.9)          & -                      & -                      & 66.7~($-$6.6)          & -                                       & -                       & 73.3 \\
      LQ-Net~\cite{zhang2018lq}             & -                       & 66.6~($-$7.2)          & 69.8~($-$4.0)          & 71.9~($-$1.9)          & -                                       & -                       & \bf{73.8} \\
      QIL~\cite{jung2019learning}           & -                       & -                      & 70.6~($-$3.1)          & 73.1~($-$0.6)          & 73.7~~\hspace{0.048cm}($+$0.0)          & -                       & 73.7 \\
      DSQ~\cite{gong2019differentiable}     & -                       & -                      & 70.0~($-$3.8)          & 72.5~($-$1.3)          & 72.8$^\dagger$~($-$1.0)                 & -                       & \bf{73.8} \\
      IR-Net~\cite{qin2020forward}          & -                       & -                      & -                      & -                      & -                                       & 70.4~($-$2.9)           & 73.3 \\
      Ours                                  &\bf{61.5}~(\bf{$-$11.8}) &\bf{69.6}~(\bf{$-$3.7}) &\bf{71.4}~(\bf{$-$1.9}) &\bf{73.3}~(\bf{$+$0.0}) &\bf{73.9}~~\hspace{0.048cm}(\bf{$+$0.6}) & \bf{72.2}~(\bf{$-$1.1}) & 73.3 \\
      %                                       1/1                       1/2                     2/2                      3/3                       4/4                                       1/32                      32/32
      \hline
   \end{tabular}
   \vspace{-0.2cm}
   \caption{Quantitative comparison of top-1 validation accuracy on ImageNet~\cite{deng2009imagenet} using the ResNet-34~\cite{he2016deep} architecture. We report results for quantized networks and their full-precision versions. $^\dagger$: all layers including the first and the last layers are quantized.}
   \vspace{-0.15cm}
   \label{tab:resnet34}
\end{table*}

\begin{table}[t]
   \setlength{\tabcolsep}{0.3em}
   \captionsetup{font={small}}
   \centering
   \small
   \vspace{-0.1cm}
   \begin{tabular}{l|C{1.9cm}||C{1.3cm}}
      \hline
      \diagbox{Methods}{W/A} & 4/4 & 32/32 \\
      \hline
      \hline
      PACT~\cite{choi2018pact,wang2019haq}  & 61.4~\hspace{0.048cm}($-$10.4)     & 71.8 \\
      DSQ~\cite{gong2019differentiable}     & 64.8~~~\hspace{0.048cm}($-$7.1)    & \bf{71.9} \\
      PROFIT~\cite{park2020profit}          & \bf{71.6}$^\dagger$~~(\bf{$-$0.3}) & \bf{71.9} \\
      Ours                                  & 70.3$^\dagger$~~($-$1.6)           & \bf{71.9} \\
      %                                       4/4                                  32/32
      \hline
   \end{tabular}
   \vspace{-0.2cm}
   \caption{Quantitative comparison of top-1 validation accuracy on ImageNet~\cite{deng2009imagenet} using the MobileNet-V2~\cite{sandler2018mobilenetv2} architecture. We quantize MobileNet-V2 using our method with the training hyperparameters and the network structures used in PROFIT~\cite{park2020profit}. We report results for quantized networks and their full-precision versions. $^\dagger$: all layers including the first and the last layers are quantized.}
   \label{tab:mobilenet}
   \vspace{-0.3cm}
\end{table}

\subsection{Results} \label{sec:exp_results}
We compare in Table~\ref{tab:resnet18} the validation accuracy on ImageNet~\cite{deng2009imagenet} using the ResNet-18~\cite{he2016deep} architecture under various bit-width settings. All numbers for other methods, except for LSQ~\cite{esser2019learned}, are taken from corresponding papers including the performance of full-precision models. Note that LSQ reports the results with a pre-activation structure~\cite{he2016identity} of ResNet, which is different from ours. We thus take the results from the work of~\cite{bhalgat2020lsq+} in which LSQ is reproduced using the same network structure as ours. We summarize the findings from Table~\ref{tab:resnet18} as follows: (1) Our quantization method with EWGS achieves the state of the art. For 4-bit weights and 4-bit activations, our method shows the classification accuracy slightly lower than LSQ+~\cite{bhalgat2020lsq+}, but the performance gain w.r.t the full-precision model is on a par with LSQ+. In particular, our model achieves the performance comparable to the full-precision one with only 3-bit representations. (2) Our method performs better than QuantNet~\cite{yang2019quantization} and DSQ~\cite{gong2019differentiable}, which attempt to address the problem of STE using soft quantizers, indicating that EWGS is a better alternative to STE than soft quantizers. (3) Our method exploiting EWGS brings significant performance improvement in a binary setting, surpassing other methods including the ones specially designed for binary quantization~\cite{rastegari2016xnor,qin2020forward}. This suggests that EWGS works favorably even with large discretization errors. (4) We verify the effectiveness of our method over a wide range of quantization bit-widths, outperforming the state of the art consistently, whereas other methods report the results selectively in few settings.

We show in Tables~\ref{tab:resnet34} and~\ref{tab:mobilenet} quantization results for ResNet-34~\cite{he2016deep} and MobileNet-V2~\cite{sandler2018mobilenetv2}, respectively, on ImageNet~\cite{deng2009imagenet}. As mentioned earlier, LSQ~\cite{esser2019learned} uses a different network structure\footnote{The full-precision baseline of ResNet-34 used in LSQ shows the top-1 validation accuracy of 74.1, which is higher than that of our full-precision baseline (73.3).}, but we include its performance in Table~\ref{tab:resnet34} in order to compare relative performance drops or gains for network quantization. We can observe from Table~\ref{tab:resnet34} similar findings in Table~\ref{tab:resnet18}. Our method outperforms the state of the art over all bit-width settings. With 3-bit weights and 3-bit activations, the quantized network trained with our method does not degrade the performance at all, compared with the full-precision model. Ours also gives better results than LSQ in terms of performance drops or gains after quantization. We can see from Table~\ref{tab:mobilenet} that our model performs better than PACT~\cite{choi2018pact,wang2019haq} and DSQ~\cite{gong2019differentiable}, but it is slightly outperformed by PROFIT~\cite{park2020profit} for 4-bit quantization of both weights and activations. Note that PROFIT exploits many training heuristics, such as knowledge distillation~\cite{hinton2015distilling}, progressive quantization~\cite{zhuang2018towards}, an exponential moving average of weights, batch normalization post-training, and iterative training with incremental weight freezing. We achieve a comparable result using a simple gradient scaling without bells and whistles, which confirms that EWGS is simple yet effective method for network quantization. Moreover, PROFIT is effective to quantize light-weight networks only, while ours can be applied to various network architectures under a wide range of bit-widths.

\subsection{Discussion} \label{sec:exp_analy}
\begin{figure}[t]
   \captionsetup{font={small}}
   \begin{center}
      \vspace{-0.2cm}
      \begin{subfigure}{1\columnwidth}
         \centering
         \includegraphics[width=0.48\columnwidth]{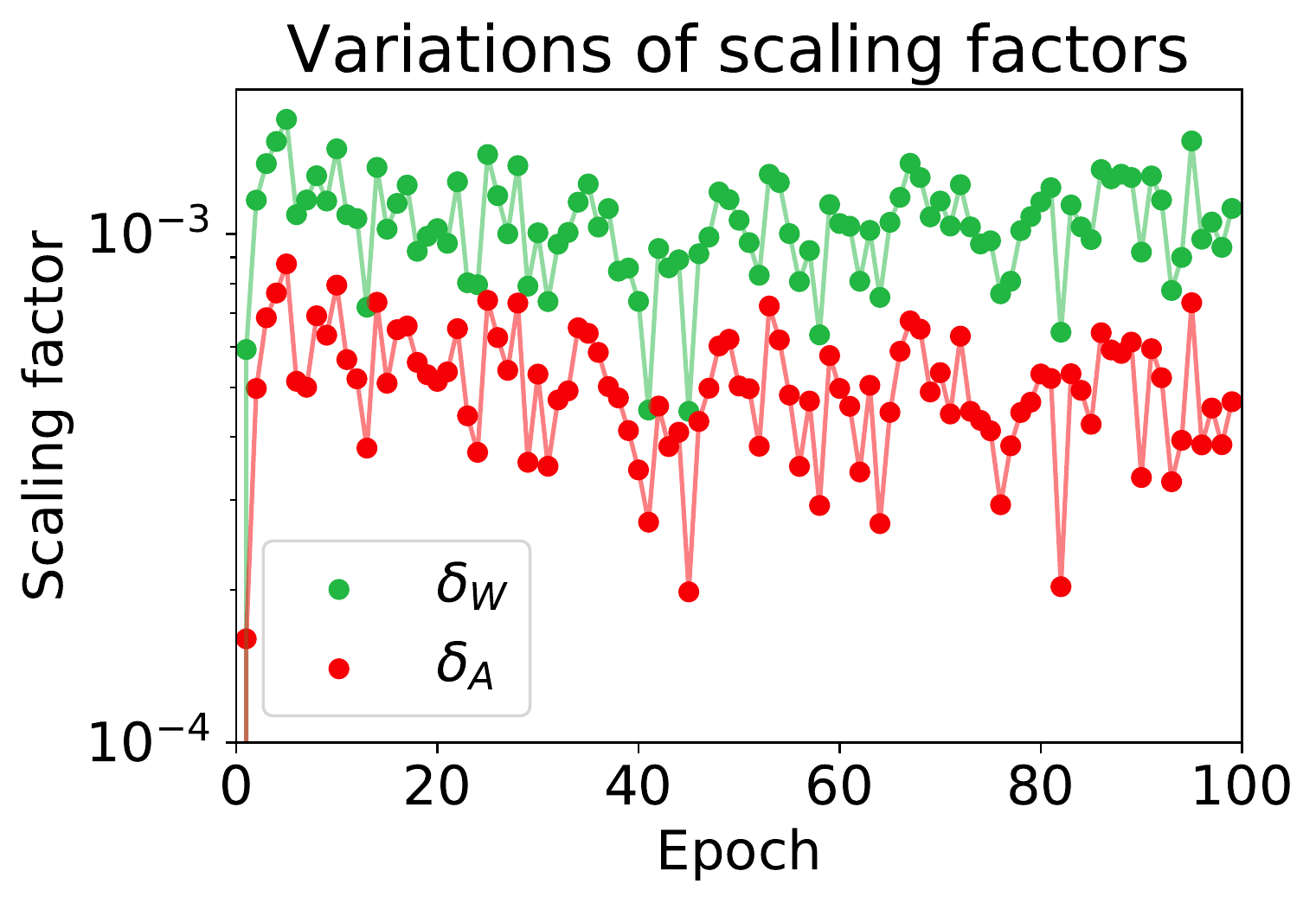}
         \includegraphics[width=0.50\columnwidth]{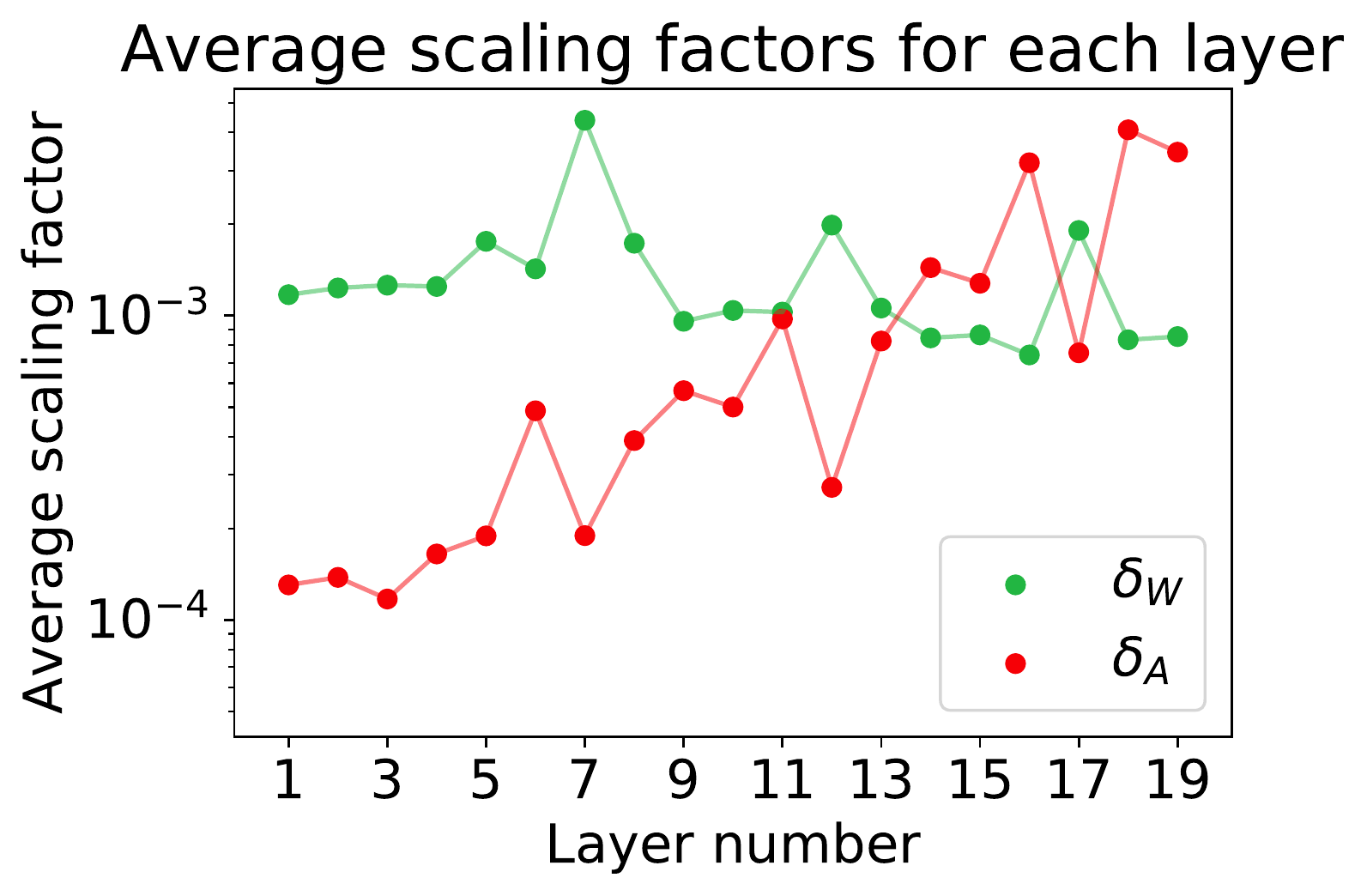}
         \vspace{-0.15cm}
      \end{subfigure}
   \end{center}
   \vspace{-0.4cm}
      \caption{Variations of scaling factors in the 10$^{\text{th}}$ quantized layer over training epochs~(left); Scaling factors for each layer averaged over epochs~(right). We visualize scaling factors for both weight and activation, denoted by~$\delta_W$ and~$\delta_A$, respectively. We use the ResNet-18~\cite{he2016deep} architecture for binary quantization. (Best viewed in color.)}
   \label{fig:scaling_factor}
   \vspace{-0.2cm}
\end{figure}

\paragraph{Analysis on scaling factor.}
We show in Fig.~\ref{fig:scaling_factor} variations of scaling factors at a particular layer during training. We can see that scaling factors oscillate within a certain range without diverging or changing drastically. This suggests that we could consider the scaling factors as hyperparameters, fixed regardless of training epochs, instead of updating them frequently. Figure~\ref{fig:scaling_factor} also shows scaling factors for each layer averaged over epochs. We can observe that scaling factors for weights and activations tend to decrease and increase, respectively, for deeper layers, except the 7$^{\text{th}}$, 12$^{\text{th}}$, and 17$^{\text{th}}$ layers. They correspond to convolutional layers with a filter size of~$1 \times 1$ and a stride of~$2$, designed to reduce the size of residuals in the residual blocks, having the different behavior compared to other plain layers. This confirms that our strategy leveraging Hessian information captures different characteristics across layers, providing an individual scaling factor for each layer.

\begin{table}[t]
   \setlength{\tabcolsep}{0.3em}
   \captionsetup{font={small}}
   \centering
   \small
   \begin{tabular}{L{4.2cm}|C{2.7cm}}
      \hline
      \multirow{2}{*}{Scaling factor} & Top-1 accuracy \\
                                    & (full-precision: 91.4) \\
      \hline
      \hline
      Eq.~\eqref{eq:update_scale} with $G = 3 \sigma (\mathcal{G}_{\mathbf{x}_{q}})$                   & \bf{85.6} \\
      Eq.~\eqref{eq:update_scale} with $G = \mathrm{max}(\vert \mathcal{G}_{\mathbf{x}_{q}} \vert)$    & 85.5 \\
      Eq.~\eqref{eq:update_scale} with $G = \mathbb{E}[\vert g_{x_q} \vert]$                           & 83.1 \\
      \hline
      Fixed~(1e-1)      & 60.9 \\
      Fixed~(1e-3)     & \bf{85.3} \\
      Fixed~(1e-5)     & 85.0 \\
      Fixed~(0) = STE  & 84.7 \\
      \hline
   \end{tabular}
   \vspace{-0.2cm}
   \caption{Quantitative comparison for different configurations of scaling factors. We binarize both weights and activations of ResNet-20~\cite{he2016deep} on CIFAR-10~\cite{krizhevsky2009learning}, and report the top-1 test accuracy. The first three rows use scaling factors obtained by Eq.~\eqref{eq:update_scale} but with different gradient representatives~$G$. The last four rows use fixed hyperparameters, specified in brackets, for both scaling factors of weights and activations in all quantized layers.}
   \label{tab:scaling_factor}
   \vspace{-0.4cm}
\end{table}

\begin{table}[t]
   \setlength{\tabcolsep}{0.15em}
   \captionsetup{font={small}}
   \centering
   \small
   \begin{tabular}{C{2.62cm}|C{0.75cm}|C{1.8cm}|C{1.4cm}|C{1.0cm}}
      \hline
      Network 			& \multirow{2}{*}{W/A}  & Quant.    	 & Backward 	& Top-1	\\
      architectures 	&                       & methods       & methods    	& acc.    \\ 
      \hline
      \hline
      \multicolumn{5}{c}{ImageNet} \\
      \hline
      \hline
      \multirow{4}{*}{ResNet-18~\cite{he2016deep}}                & \multirow{2}{*}{1/1}  & \multirow{2}{*}{Ours}                          & STE    & 54.6 \\
                                                                  &                       &                                                & EWGS   & \bf{55.3} \\\cline{2-5}
                                                                  & \multirow{2}{*}{1/32} & \multirow{2}{*}{Ours}                          & STE    & 66.3 \\
                                                                  &                       &                                                & EWGS   & \bf{67.3} \\
      \hline
      \multirow{4}{*}{MobileNet-V2~\cite{sandler2018mobilenetv2}} & \multirow{2}{*}{4/4$^\dagger$}  & \multirow{2}{*}{Ours}                          & STE             & 69.2 \\
                                                                  &                                 &                                                & EWGS            & \bf{70.3} \\\cline{2-5}
                                                                  & \multirow{2}{*}{4/4$^\dagger$}  & \multirow{2}{*}{PROFIT~\cite{park2020profit}}  & STE             & ~\hspace{0.058cm}69.2$^\star$ \\
                                                                  &                                 &                                                & EWGS$^\ddagger$ & \bf{70.0} \\
      \hline
      \hline
      \multicolumn{5}{c}{CIFAR-10} \\
      \hline
      \hline
      \multirow{6}{*}{ResNet-20~\cite{he2016deep}}                & \multirow{2}{*}{1/1}            & \multirow{2}{*}{Ours}                          & STE                            & 84.7 \\
                                                                  &                                 &                                                & EWGS                           & \bf{85.6} \\\cline{2-5}
                                                                  & \multirow{2}{*}{1/1}            & \multirow{2}{*}{DoReFa~\cite{zhou2016dorefa}}  & STE                            & ~\hspace{0.058cm}84.9$^\star$ \\
                                                                  &                                 &                                                & EWGS$^\ddagger$                & \bf{85.9} \\\cline{2-5}
                                                                  & \multirow{2}{*}{1/32$^\dagger$} & \multirow{2}{*}{DoReFa~\cite{zhou2016dorefa}}  & STE                            & ~\hspace{0.058cm}89.7$^\star$ \\
                                                                  % &                                 &                                                & MetaQ~\cite{chen2019metaquant} & 90.0 \\
                                                                  &                                 &                                                & EWGS$^\ddagger$                & \bf{90.3} \\
      \hline
   \end{tabular}
   \vspace{-0.25cm}
   \caption{Quantitative comparison of STE and EWGS. We use ResNet-18~\cite{he2016deep} and MobileNet-V2~\cite{sandler2018mobilenetv2} on ImageNet~\cite{deng2009imagenet}, and ResNet-20 on CIFAR-10~\cite{krizhevsky2009learning}. We report the top-1 validation and test accuracies for ImageNet and CIFAR-10, respectively. For MobileNet-V2 and ResNet-20, we also compare the performance with different quantization methods, such as PROFIT~\cite{park2020profit} and DoReFa-Net~\cite{zhou2016dorefa}. $^\dagger$: all layers including the first and the last layers are quantized; $^\ddagger$: weight and activation scaling factors for EWGS in all quantized layers are fixed to~$0.01$; $^\star$: models reproduced by ourselves.}
   \label{tab:ours_vs_ste}
   \vspace{-0.4cm}
\end{table} 

We compare quantization results for different configurations of scaling factors in Table~\ref{tab:scaling_factor}. We obtain the results by binarizing weights and activations of ResNet-20~\cite{he2016deep} on CIFAR-10~\cite{krizhevsky2009learning}. The first three rows show quantization results for different gradient representatives~$G$ in Eq.~\eqref{eq:update_scale}. Overall, our method shows better results with large gradient elements,~\eg.,~three standard deviations and the maximum over absolute values in the first and second rows, respectively, than the small one,~\eg.,~an average in the third row. A reason is that large gradients mainly influence the training process, but an average value is usually biased to small gradient elements, as discussed in Sec.~\ref{sec:scaling_factor}. The last four rows compare the results with fixed scaling factors. We use the same scaling factor for both weight and activation quantizers in all quantized layers. We can see that our method achieves the performance comparable to the best result~(85.6 vs. 85.3), and even outperforms STE~(84.7 vs. 85.3), if the scaling factor is properly set. Otherwise, the performance is degraded~(\eg,~with the scaling factor of~1e-1) or becomes similar to the one for STE, especially with an extremely small scaling factor~(\eg,~1e-5). This suggests that EWGS is also effective with a single scaling factor, but the value should be carefully chosen.
\vspace{-0.4cm}

\begin{figure}[t]
   \captionsetup{font={small}}
   \begin{center}
      \begin{subfigure}{1\columnwidth}
         \centering
         \includegraphics[width=0.495\columnwidth]{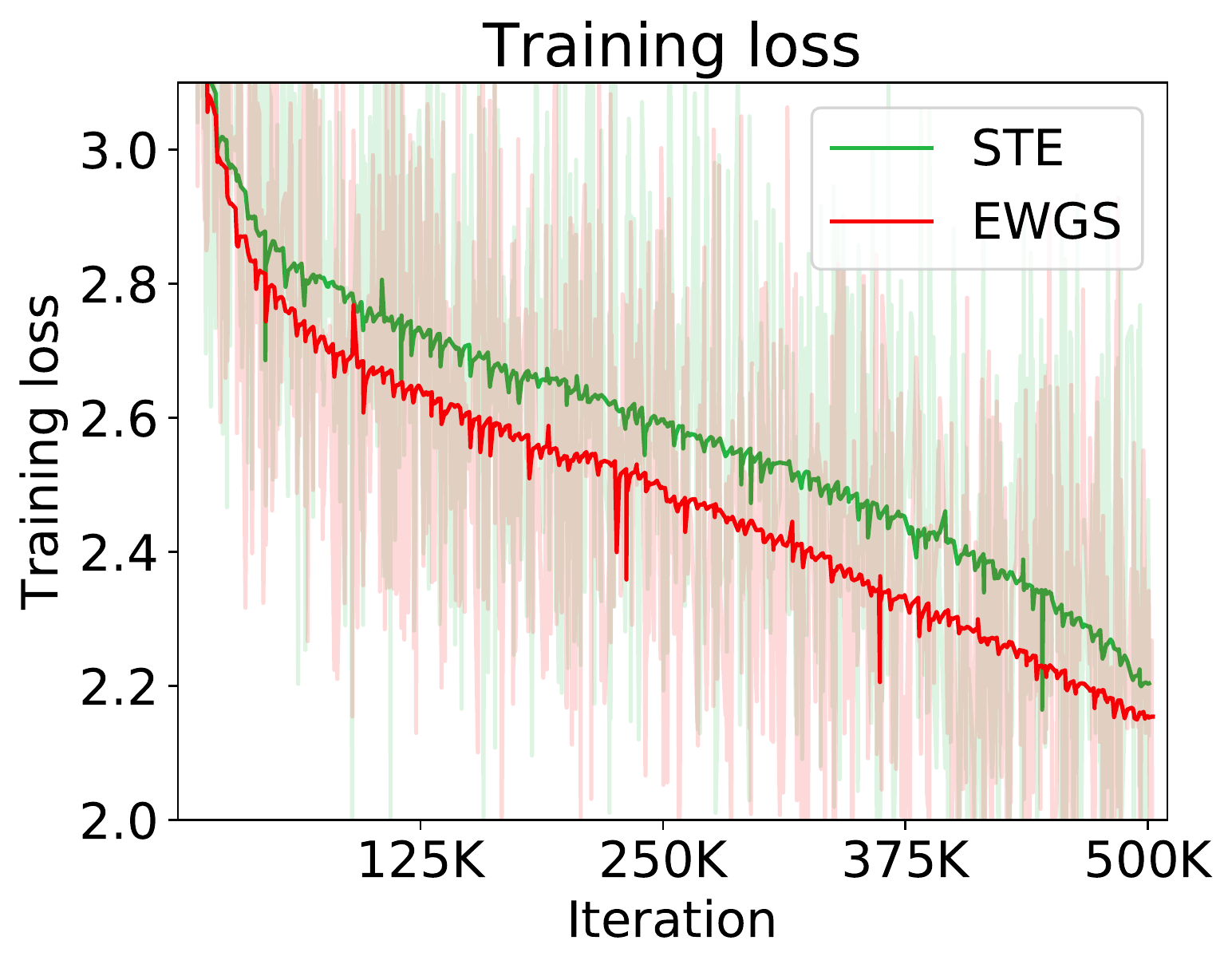}
         \includegraphics[width=0.485\columnwidth]{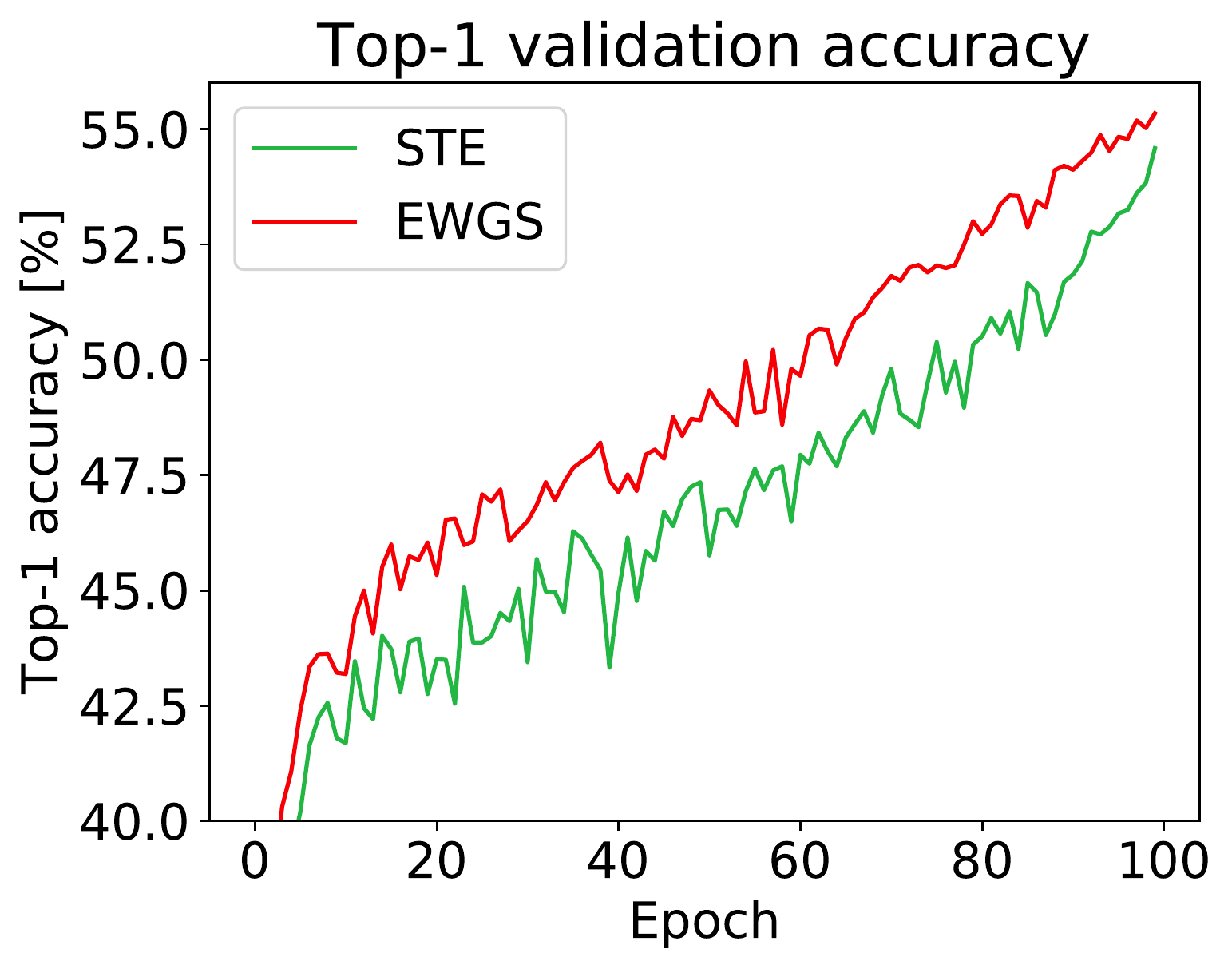}
         \vspace{-0.15cm}
         \caption{Weight: 1-bit / Activation: 1-bit.}
         \label{fig:W1A1_curves}
      \end{subfigure}
      \begin{subfigure}{1\columnwidth}
         \centering
         \includegraphics[width=0.505\columnwidth]{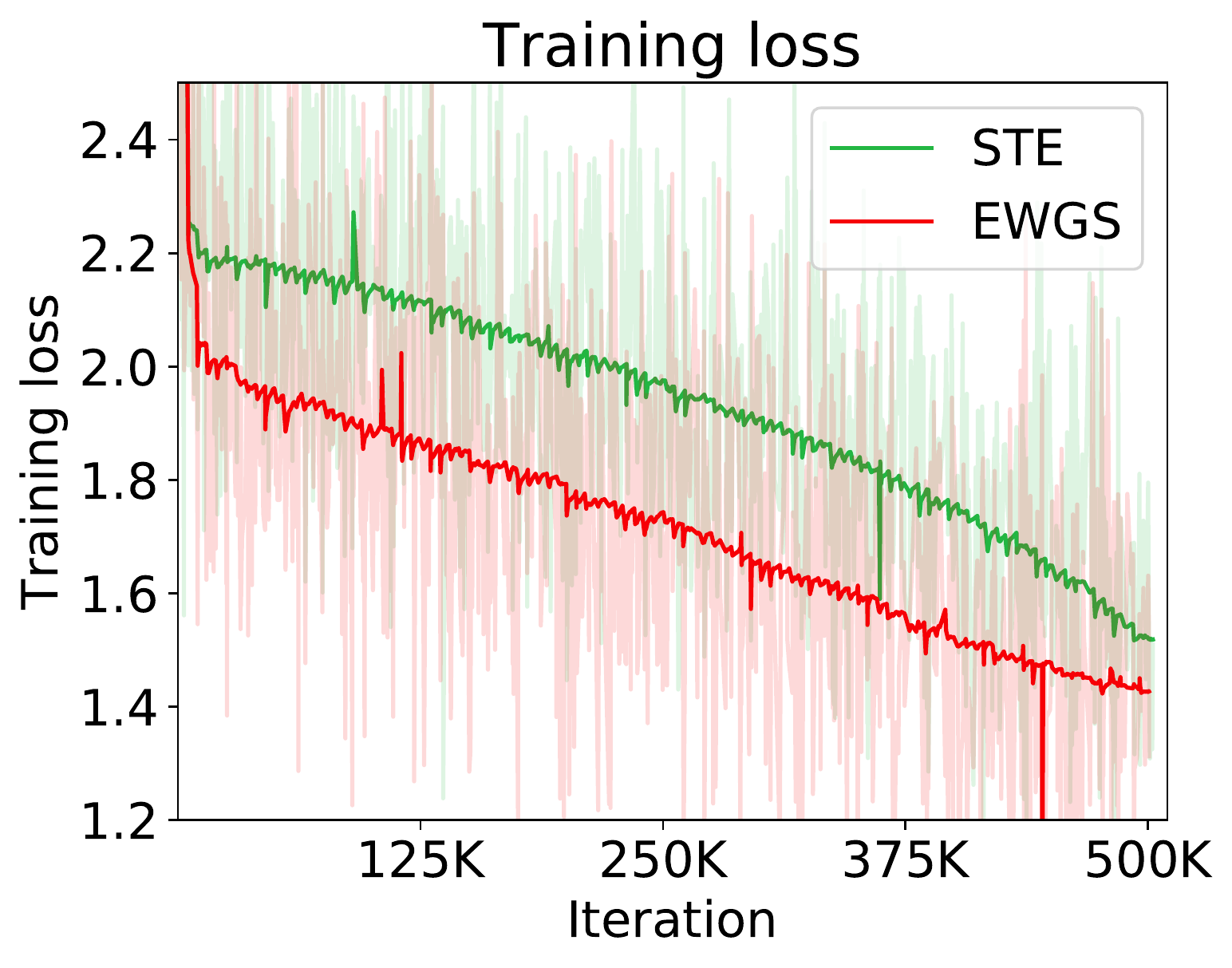}
         \includegraphics[width=0.475\columnwidth]{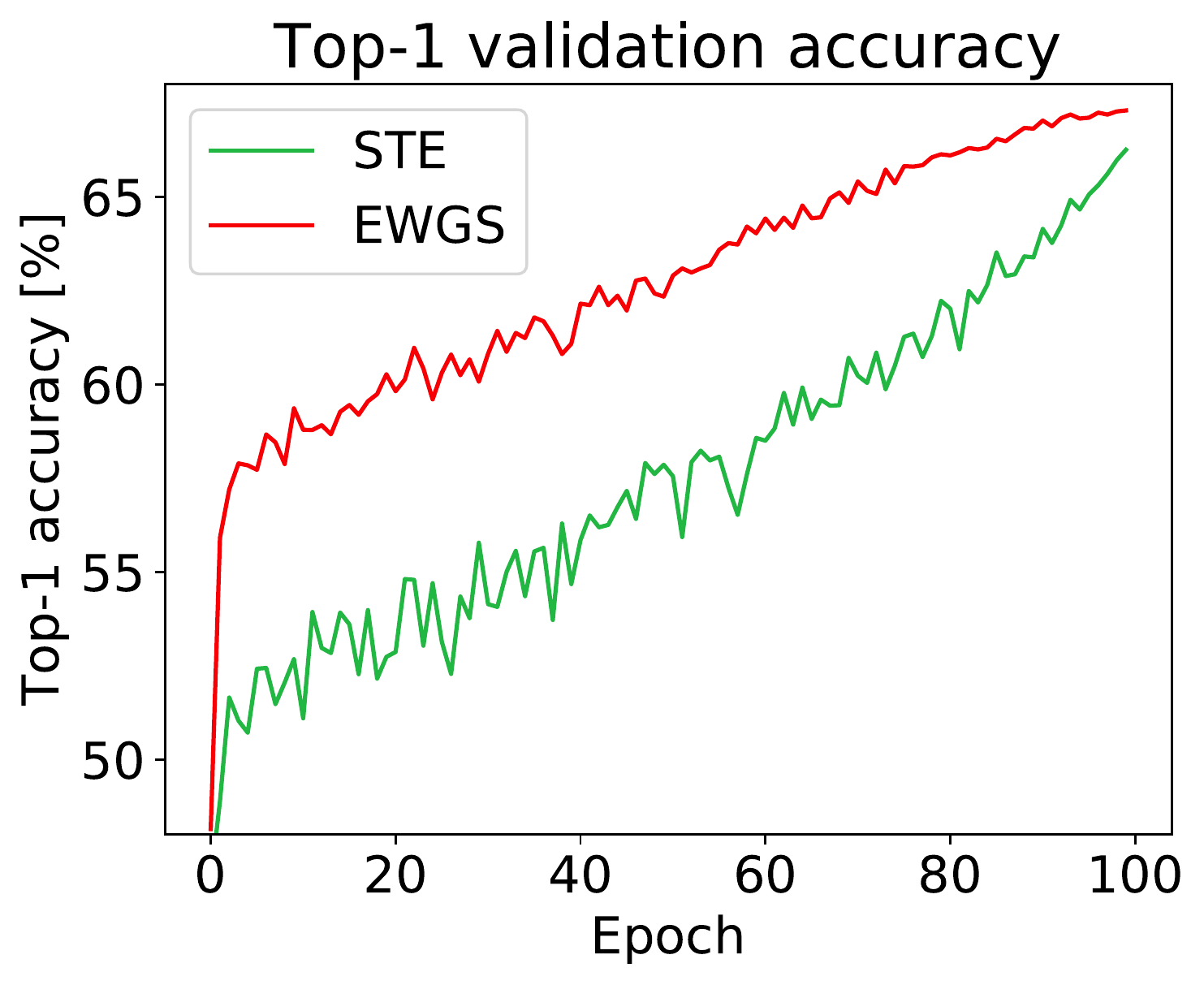}
         \vspace{-0.15cm}
         \caption{Weight: 1-bit / Activation: 32-bit.}
         \label{fig:W1A32_curves}
      \end{subfigure}
   \end{center}
   \vspace{-0.6cm}
      \caption{Training losses and validation accuracies for binarized networks using STE and EWGS. We use ResNet-18~\cite{he2016deep} to quantize (a) both weights and activations and (b) weights only, and show the results on ImageNet. (Best viewed in color.)}
   \label{fig:training_curves}
   \vspace{-0.3cm}
\end{figure}

\paragraph{Performance comparison with STE.} 
We compare in Table~\ref{tab:ours_vs_ste} the performance of EWGS and STE with different combinations of network architectures, quantization methods, and bit-widths. Specifically, we use different quantization methods, including PROFIT~\cite{park2020profit}, DoReFa-Net~\cite{zhou2016dorefa}, and ours, and exploit either EWGS or STE for backpropagation. We then use them to quantize ResNet-18~\cite{he2016deep}, MobilNet-V2~\cite{sandler2018mobilenetv2} and ResNet-20. EWGS gives better results than STE within our framework, achieving about 1\% accuracy gains over STE consistently, regardless of the network architectures. It also outperforms STE by a large margin for other quantization methods, such as PROFIT~\cite{park2020profit} and DoReFa-Net~\cite{zhou2016dorefa}, demonstrating the generalization ability of EWGS. The accuracy of PROFIT is slightly lower than the one reported in the paper, possibly because we do not use the progressive quantization technique~\cite{zhuang2018towards}. We show in Fig.~\ref{fig:training_curves} the training losses and validation accuracies for binarizing the ResNet-18~\cite{he2016deep} architecture in Table~\ref{tab:ours_vs_ste}. We can clearly see that training quantized networks with EWGS is better in terms of stability and accuracy, compared to STE. The networks with EWGS achieve lower losses and higher accuracies, which is significant especially for weight-only quantization. These results confirm once more the effectiveness of EWGS.

\vspace{-0.125cm}
\section{Conclusion}
\vspace{-0.125cm}
We have introduced an EWGS method that adjusts gradients and scaling factors adaptively for each layer. The various CNN architectures quantized by our method show state-of-the-art results for a wide range of bit-widths. We have shown that EWGS boosts the quantization performance of other methods exploiting STE, without bells and whistles, demonstrating the effectiveness and generalization ability of our approach to scaling gradients adaptively for backpropagation. We believe that EWGS could be an effective alternative to STE for network quantization.

\vspace{-0.3cm}
\paragraph{Acknowledgments.}
This research was supported by the Samsung Research Funding \& Incubation Center for Future Technology (SRFC-IT1802-06).

{\small
\bibliographystyle{ieee_fullname}
\bibliography{egbib}
}

\end{document}